\documentclass[runningheads]{llncs}
\usepackage{graphicx}
\usepackage{amsmath}
\usepackage{booktabs}
\usepackage{colortbl}
\usepackage{algpseudocode}
\usepackage{algorithm} 
\usepackage{adjustbox}
\usepackage{color}
\usepackage{amsfonts}
\usepackage{amssymb}
\usepackage{bbding}
\usepackage{multirow}
\begin{document}
%
\title{Switchable Self-attention Module}
%
%

\author{Shanshan Zhong, Wushao Wen\thanks{Corresponding authour}, Jinghui Qin}
\institute{School of Computer Science and Engineering \\Sun Yat-sen University}

%
\maketitle              
\begin{abstract}
Attention mechanism has gained great success in vision recognition. Many works are devoted to improving the effectiveness of attention mechanism, which finely design the structure of the attention operator. These works need lots of experiments to pick out the optimal settings when scenarios change, which consumes a lot of time and computational resources. In addition, a neural network often contains many network layers, and most studies often use the same attention module to enhance different network layers, which hinders the further improvement of the performance of the self-attention mechanism. To address the above problems, we propose a self-attention module SEM. Based on the input information of the attention module and alternative attention operators, SEM can automatically decide to select and integrate attention operators to compute attention maps. The effectiveness of SEM is demonstrated by extensive experiments on widely used benchmark datasets and popular self-attention networks.

\end{abstract}
\section{Introduction}
Attention is a recognition mechanism which is capable of ignoring non-essential information and selectively focusing on a small subset of information~\cite{anderson2005cognitive}. Attention is widely used in sentences~\cite{he2021blending,huang2020efficient,wang2021s2san}, images~\cite{huang2022lottery,zhang2019self,yang2017neural,wang2020hierarchical}, and videos~\cite{ma2002user,zhai2006visual,li2019beyond} to relieve the pressure of neural networks to learn massive amounts of information. Especially in vision recognition, some operators~\cite{hu2018squeeze,woo2018cbam,qin2021fcanet} imitating attention mechanism serve as effective feature enhancement components in convolutional neural networks(CNNs), enabling deep neural networks to effectively identify important information in images. Such modular and pluggable operators of neural networks are called attention modules~\cite{hu2018squeeze,woo2018cbam,park2018bam,wang2018non}, which facilitate the development of vision recognition.

\begin{figure}[t]
  \centering
  \includegraphics[width=0.6\linewidth]{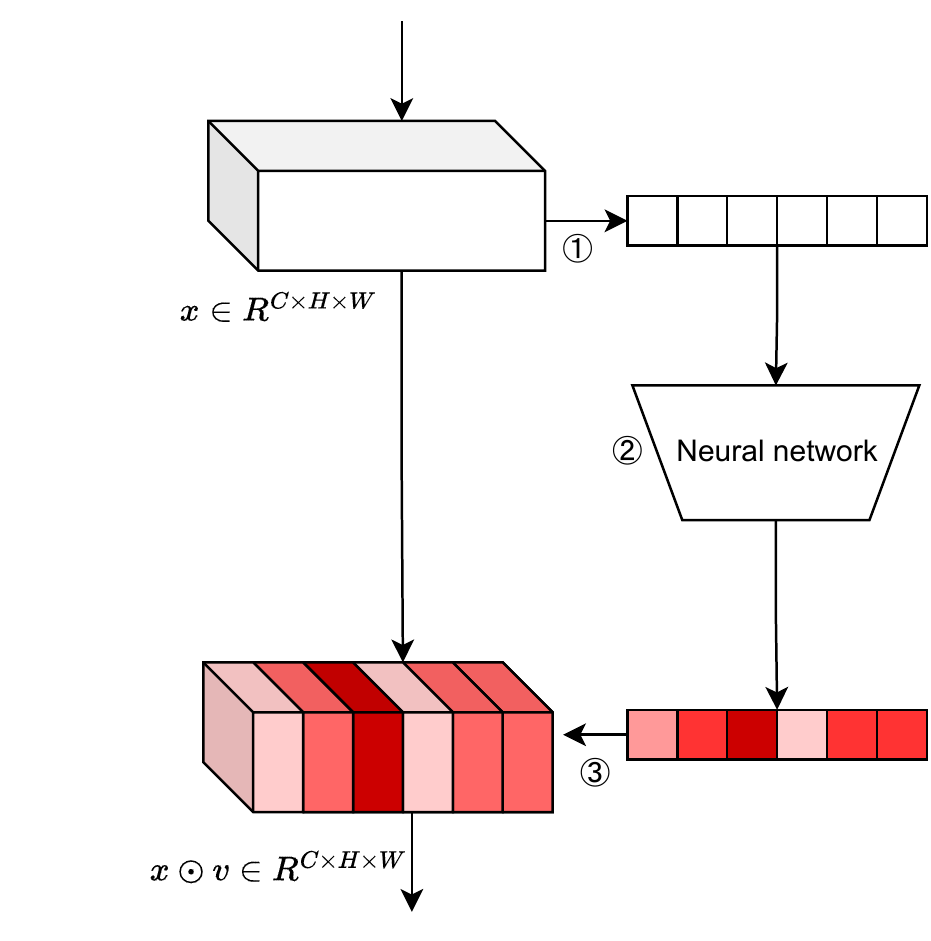}
  \caption{Diagram of each attention sub-module. the attention module can be divided into three stages~\cite{hu2018squeeze,huang2020dianet}: \raisebox{.9pt}{\textcircled{\raisebox{-.9pt}{1}}} squeeze, \raisebox{.9pt}{\textcircled{\raisebox{-.9pt}{2}}} excitation; and \raisebox{.9pt}{\textcircled{\raisebox{-.9pt}{3}}} recalibration.}
\label{fig:attention}
\end{figure}

Specifically, most of the existing attention modules mainly consist of three parts~\cite{woo2018cbam,huang2020dianet}. As shown in Fig.~\ref{fig:attention}, \raisebox{.9pt}{\textcircled{\raisebox{-.9pt}{1}}} is the squeeze module, in which we get global information embeddings from input feature maps. \raisebox{.9pt} {\textcircled{\raisebox{-.9pt}{2}}} is the excitation module, where the global information embeddings is passed through the excitation operator to extract the attention maps. Finally, in stage \raisebox{.9pt}{\textcircled{\raisebox{-.9pt}{3}}} named as recalibration, the attention maps are used to the layers and adjust the feature.

\raisebox{.9pt}{\textcircled{\raisebox{-.9pt}{2}}} is a core part of attention modules. Researchers have proposed many methods to optimize the excitation operator~\cite{hu2018squeeze,qin2021fcanet,woo2018cbam,huang2020dianet,2020ECA}. For example, SENet~\cite{hu2018squeeze} uses a fully connected network(FC) to fully capture channel-wise dependencies, which adaptively recalibrates channel-wise feature responses by explicitly modelling interdependencies between channels; ECA~\cite{2020ECA} is a local crosschannel interaction strategy without dimensionality reduction, which can be efficiently implemented via an 1D convolutional neural network(CNN); IEBN~\cite{liang2020instance} is an attention-based BN that recalibrates the information of each channel by a simple linear transformation named instance enhance(IE). Despite their great successes, existing practices often employ the same kind of attention modules in all layers of an entire CNN with suboptimal performance. These approaches ignore two important things. On the one hand, CNN is a layered feature extractor often containing many network layers, and usually the attention modules of these network layers all use the same kind of excitation operators. On the other hand, for different scenarios (inputs, datasets and tasks, etc.), the size of the inputs and the type of datasets are different and we can only select a suitable attention module based on exploratory experiments.

\begin{figure}[t]
  \centering
  \includegraphics[width=\linewidth]{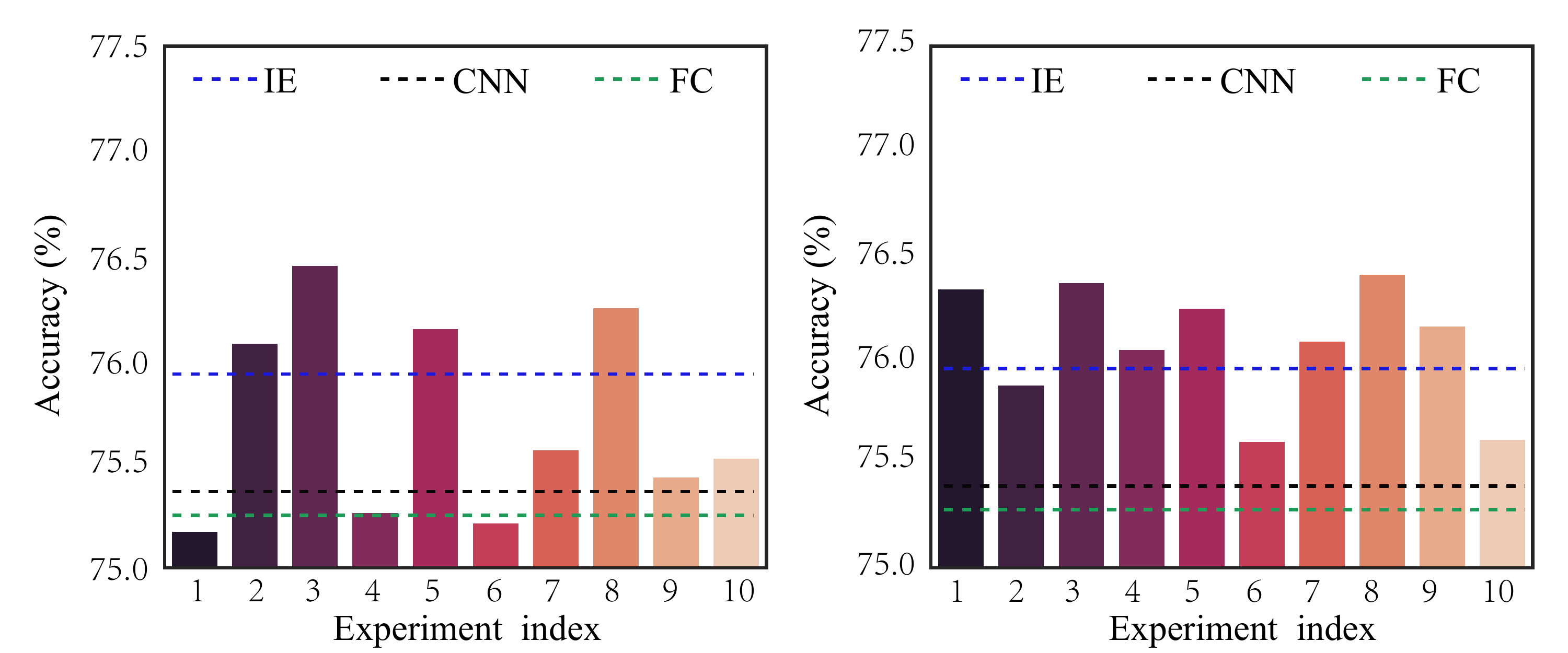}
  \caption{We conduct random experiments which includes selecting simple and double excitation operators. Left: we select a excitation operator for different layers randomly, which shows that it is efficient to randomly select different operators for different layers. Right: a random combination of double operators reveals that it is efficient to randomly select different operator combination for different layers and the performance of combination is better than simple operator.}
\label{fig:random}
\end{figure}

However, based on our exploratory experiments, we find that it is necessary to select appropriate excitation operators for different network layers and scenarios, and combining different excitation operators can improve the effectiveness of attention modules. On CIFAR100, we randomly select excitation operators for different network layers. The experimental results are shown in Fig.~\ref{fig:random}. Through multiple groups of random repeated experiments, we dig out that attention modules of different network layers may require different excitation operators. In addition, we randomly reorganize the excitation operators of different network layers in pairs, and get that a combination of excitation operators can improve the performance of attention modules, which is better than randomly selecting a simple operator. Therefore, when we use attention modules to enhance the capability of the feature maps of CNNs, ideally we need to adjust the type of excitation operators according to the network layers and scenarios, instead of using the same kind of attention modules all the way. However, choosing the appropriate excitation operator for different network layers and scenarios requires a lot of cost and time.

In order to solve the above problems, we propose a switchable excitation module (SEM), which can automatically decide to select and integrate attention operators to compute the attention maps, realizing the combination of different excitation operators in different network layers. The contributions of this paper can be summarized as follows:
\begin{itemize}
    \item We find that different network layers and scenarios require different excitation operators.
    \item In response to the above findings, we propose a switchable excitation module (SEM). SEM can select appropriate and integrate excitation operators for different network layers and scenarios to improve the excitation module.
    \item The extensive experiments of SEM show that our method can achieve the state-of-the-art results on several popular benchmarks.
\end{itemize}

\section{Related Works}

\noindent\textbf{Attention Mechanism in CNNs.}
CNNs with attention mechanism are widely used in a variety of vision tasks. In recent years, Zhu et al.~\cite{2019Evolutionary} has explored the neuro-evolution application to the automatic design of CNN topologies, developing a novel solution based on Artificial Bee Colony. Gao et al.\cite{gao2019conditional} proposes a novel CRF layer for graph convolutional neural networks to encourage similar nodes to have similar hidden features. Besides these works, many reseachers try to extend the attention mechanisms to specific tasks, e.g. point cloud classification \cite{xie2018attentional,huang2021rethinking}, image super-resolution \cite{zhang2018image}, object detection \cite{dai2017deformable,huang2020convolution}, semantic segmentation \cite{yuan2018ocnet,2020Dual}, face recognition \cite{yang2017neural}, person re-identification \cite{li2018harmonious}, action recognition \cite{wang2018non}, image generation \cite{gregor2015draw,zhang2019self}, and 3D vision \cite{xie2018attentional}.

\noindent\textbf{Excitation Module of Attention Mechanism.}There are many works to improve the capability of excitation module. SENet~\cite{hu2018squeeze} adaptively recalibrates channel-wise feature responses by explicitly modelling interdependencies between channels, which makes use of fully connected networks. LightNL~\cite{gao2019global} squeezes the transformation operations and incorporating compact features. ECA~\cite{2020ECA} can be efficiently implemented via 1D convolution. IEBN~\cite{liang2020instance} recalibrates the information of each channel by a simple linear transformation.  CAP~\cite{behera2021context} effectively captures subtle changes via sub-pixel gradients which cognizes subcategories by introducing a simple formulation of context-aware attention via learning where to look when pooling features across an image. Unlike these studies that focus on improving the performance of certain aspects of excitation operators, SEM can automatically decide to select and integrate attention operators to compute the attention maps, realizing the combination of different excitation operators in different network layers.

\section{Methodology}
In this section, we formally introduce SEM. We first introduce the general structure of the SEM. Later, we introduce the details of decision module and switching module which are specific of SEM.

\begin{algorithm}[t]  
    \caption{The algorithm of producing attention map from SEM}
    \label{alg:ean}   
    \textbf{Input:}  A feature map $x\in R^{C\times H\times W}$; A learnable transformation $F$; A set of excitation operators $\text{EO}$. 
    
    \textbf{Output:} The attention map $v$.
        
    \begin{algorithmic}[1]

    \State\algorithmiccomment{Squeeze module}

    \State Calculate the global information embedding $m \gets GAP(x)$
    
    \State\algorithmiccomment{Decision module}       
    \State $w \gets F(m)$
       
    \State\algorithmiccomment{Switching module} 

    \For{$i$ from 1 to $N$} 
        \State $v_{e_i} = e_i(m)$
    \EndFor
    \State Calculate $v$ by Eq.~(\ref{eq:switch})
    
    \State\Return $v$
    \end{algorithmic} 
\end{algorithm}  

\subsection{An Overview of SEM}

\begin{figure}[t]
  \centering
  \includegraphics[width=0.8\linewidth]{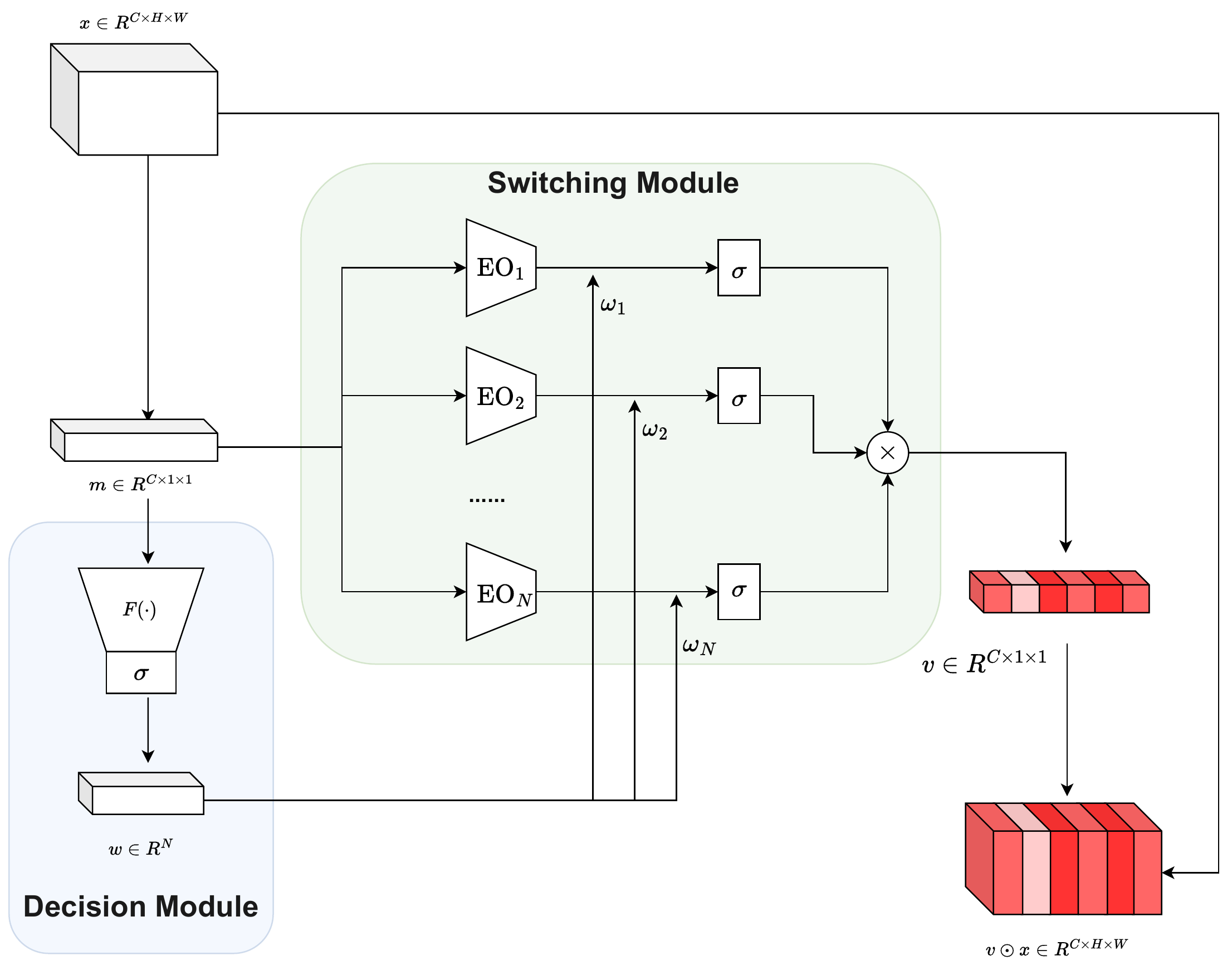}
  \caption{An illustration of SEM. We conduct inputs and outputs dimensions of different modules, as well as the computational flow. Decision module is used to generate a decision vector $w$, and switching module selects excitation operators and integrates them based on $w$ to calculate the attention feature map $v$. }
\label{fig:arch}
\end{figure}

As shown in Fig.~\ref{fig:arch}, SEM mainly improves the excitation module of attention modules, and its specific structure can be divided into two submodules: decision module and switching module. The feature map of the current network layer is defined as $x \in R^{C \times H \times W}$, and the calculation process of the attention value of $x$ is as follows. First, a global average pooling denoted as GAP($\cdot$) is applied to extract global information from features in the squeeze module as shown in Eq.(\ref{eq:squeeze}).
\begin{equation}
m = \textbf{GAP}(x),
\label{eq:squeeze}
\end{equation}
where $m \in R^{C \times 1 \times 1}$ is the global information embedding. Then, we use $m$ as the input of decision module and switching module respectively.

A decision network is designed in the decision module to generate a decision vector $w \in R^N$ for selecting excitation operators. Based on the decision vector $w$, we design the switching module to select and integrate excitation operators. The switching module includes a set of excitation operators $\text{EO}$. The size of $\text{EO}$ is defined as $N$, and $eo_i, i \in (0, N]$ is the element of $\text{EO}$, representing the alternative excitation operator. The switching module designs the strategy to calculate the attention feature map $v \in R^{C \times 1 \times 1}$ based on $w$.

Then, through the attention feature map $v$, we calculate the feature map $x_{\textbf{att}} \in R^{C \times H \times W}$ by integrating the attention information through Eq.(\ref{eq:recalibration}). 
\begin{equation}
x_{\textbf{att}} = x \odot v,
\label{eq:recalibration}
\end{equation}
where $\odot$ is dot product. We explain the details of each module below.

\subsection{Decision Module}
In order to automatically select the appropriate excitation operators for different network layers and scenarios, we design a decision module, which can generate decision weight $w$ for selecting excitation operators.

We take the global information embedding $m$ as the input, which enables the decision module to adaptively generate different decision weights according to different inputs. In addition, the decision module for each network layer is separate so that the decision module can distinguish different network layers. To make use of the information aggregated in the squeeze operation to identify the importance of different operatos, we follow it with $F(\cdot)$ which aims to fully capture decision information from channel-wise dependencies. To fulfil this objective, the function must meet two criteria: first, it must be flexible (in particular, it must be capable of learning a nonlinear interaction between channels) and second, it must learn a non-mutually-exclusive relationship since we would like to ensure that multiple excitation operator are allowed to be em phasised opposed to one-hot activation. To meet these criteria, we opt to employ a simple gating mechanism with a sigmoid activation:
\begin{equation}
w = \sigma(F(m)) = \sigma(W_d m),
\label{eq:decision}
\end{equation}
where $\sigma$ is the Sigmoid function. $F(\cdot)$ is the decision function. We use the fully connected network to define $F(\cdot)$, $W_d \in R^{N \times C}$ is the weight of the fully connected network. Each value of decision vector $w \in R^N$ represents the importance of the corresponding excitation operator. Through $w$, the decision module makes soft decisions instead of selecting a single excitation operator. This design enables the excitation module to obtain an attention map that integrates all excitation operator information according to the importance of different excitation operators.

\subsection{Switching Module}
Based on the decision vector $w$, we design the switching mechanism to select and integrate excitation operators. In the experimental exploration of this paper, we set the value of $N$ to 3, and set FC, CNN and IE as alternative excitation operators to explore the structure of the switching module.

\noindent\textbf{Fully Connected Neural Networks. }We use the excitation operator of SENet~\cite{hu2018squeeze} as the full connected neural network. SENet is a classic fully connected network based attention module, which parameterises the gating mechanism by forming a bottleneck with two fully connected layers around the non-linearity to limit model complexity and aid generalisation. We opt to employ SENet as shown in Eq.(\ref{eq:fc}).
\begin{equation}
v_{\textbf{fc}} = F_{ex}(m, W) = g(m, W) = W_2\delta(W_1m),
\label{eq:fc}
\end{equation}
where $\delta$ refers to the ReLU function, $W_1 \in R^{\frac{C}{r} \times C}$ and $W_2 \in R^{C \times \frac{C}{r} }$. $r$ is the reduction ratio.

\noindent\textbf{Convolutional Neural Networks. }CNNs have the characteristics of few parameters and simple network structure, which can be well used to extract image features. There are many studies~\cite{2020ECA,fu2017look,pan2022integration} using CNNs as the excitation operator of the attention module. We choose a simple-yet-effective attention module ECA~\cite{2020ECA} as the excitation operator of the CNN type. ECA only involves a handful of parameters while determining clear performance gain, which is a method to adaptively select kernel size of 1D convolution, coverage of local cross-channel interaction. We consider a standard convolution with the kernel $K \in R^{ C \times C \times k \times k}$, where $k$ is the kernel size and $C$ are the channel size. Given tensors $m \in R^{C \times 1 \times 1}$, $ v_{cnn} \in R^{C \times 1 \times 1}$ as the input and output feature maps, we denote $m_{ij} \in R^C$, $v_{ij} \in R^ C$ as the feature tensors of pixel ($i, j$) corresponding to $m$ and $v_{cnn}$ respectively. Then, the standard convolution can be formulated as:
\begin{equation}
v_{ij} = \sum_{p, q} K_{p, q} m_{i+p-\lfloor k/2 \rfloor, j+q-\lfloor k/2 \rfloor},
\label{eq:cnn}
\end{equation}
where $K_{p,q} \in R^{C \times C} , p, q \in {0, 1, \cdot \cdot \cdot , k-1}$, represents the kernel weights with regard to the indices of the kernel position ($p, q$).

The value of $k$ is canculate based on the value of $C$~\cite{2020ECA}.
\begin{equation}
k = \psi(C) = | \frac{log_2(C)}{\gamma} + \frac{b}{\gamma}|_{odd},
\label{eq:k}
\end{equation}
where $|t|_{odd}$ indicates the nearest odd number of $t$. We set $\gamma$ and $b$ to 2 and 1 throughout all the experiments, respectively.

\noindent\textbf{Instance Enhance. }In addition to FC and CNNs, we also consider the application of linear relationships in attention modules. We refer to the structure of IEBN~\cite{liang2020instance} and define a pair of learnable parameters $\gamma$, $\beta$ scale and shift the global information embedding $m$ to restore the representation power. We define this structure as Instance Enhance (IE). The structure of IE is very simple and it is calculated as follows~\cite{liang2020instance}:
\begin{equation}
v_{\textbf{ie}} = m \times \gamma + \beta,
\label{eq:ie}
\end{equation}
Specially, the parameters $\gamma$, $\beta$ are initialized by constant 0 and -1 espectively.

\noindent\textbf{Switching Operator. } We use $w$ to adjust the proportion of each operator in $\text{EO}$, and combine the results of each operator in the form of dot product to get the final attention feature map $v$.
\begin{equation}
v = \sigma(v_{\textbf{fc}} w_{\textbf{fc}}) \odot \sigma(v_{\textbf{cnn}} w_{\textbf{cnn}}) \odot \sigma(v_{\textbf{ie}} w_{\textbf{ie}}),
\label{eq:switch}
\end{equation}
where $\sigma$ is the sigmoid function, $w_{\ast}$ is the weight of each excitation operator.

\begin{table}[htbp]
  \centering
  \caption{Comparison of different attention modules on CIFAR10 and CIFAR100 in terms of Top-1 accuracy (in \%). }
    \begin{tabular}{l|cc}
    \toprule
    Model & CIFAR10 & CIFAR100 \\
    \midrule
    ResNet164 & 93.23 & 74.33 \\
    ResNet164+SE & 94.32 & 75.28 \\
    ResNet164+CBAM & 92.67 & 74.54 \\
    ResNet164+SRM & 94.58 & 76.11 \\
    ResNet164+ECA & 94.51 & 75.39 \\
    ResNet164+IE &  94.44 & 75.92 \\
    ResNet164+SEM(ours) & \textbf{94.95({$\uparrow$ 1.72})} & \textbf{76.76({$\uparrow$ 2.43})} \\
    \bottomrule
    \end{tabular}%
  \label{tab:main-ex}%
\end{table}%

\section{Experiments}

\noindent\textbf{Dataset and Implementation Details. }We evaluate our SEM on both CIFAR10 and CIFAR100~\cite{krizhevsky2009learning}, which have 50k train images and 10k test images of size 32 by 32 but has 10 and 100 classes respectively. We also use normalization and standard data augmentation including random cropping and horizontal flipping during training. SGD optimizer with a momentum of 0.9 and a weight decay of $1e^{-4}$ is applied in our experiments. We train all of models using one Nvidia RTX 3080 GPU and uniformly set the epoch number to 164.

\noindent\textbf{Image Classification. } 
We compare our SEM with several popular attention methods using ResNet-164 on CIFAR10 and CIFAR100, including SENet~\cite{hu2018squeeze}, CBAM~\cite{woo2018cbam}, SRM~\cite{lee2019srm}, ECA~\cite{2020ECA}, and IE~\cite{liang2020instance}. The evaluation metric is Top-1 accuracy(in \%). The results are given in Table~\ref{tab:main-ex}, where we can see that our SEM achieves 1.72\% and 2.43\% gains in Top-1 accuracy on CIFAR10 and CIFAR100 respectively. 

\begin{table}[htbp]
  \centering
  \caption{Comparison of different depth of backbone models on CIFAR10 and CIFAR100 in terms of Top-1 accuracy (in \%). }
    \begin{tabular}{l|cc}
    \toprule
    Model & CIFAR10 & CIFAR100 \\
    \midrule
    ResNet47 &   93.54    & 72.56 \\
    ResNet47+SEM(ours) &   \textbf{93.6({$\uparrow$ 0.06})}    & \textbf{73.09({$\uparrow$ 0.53})} \\
    \midrule
    ResNet164 & 93.23 & 74.33 \\
    ResNet164+SEM(ours) & \textbf{94.95({$\uparrow$ 1.72})} & \textbf{76.76({$\uparrow$ 2.43})} \\
    \midrule
    ResNet272 &   93.99   &  74.08\\
    ResNet272+SEM(ours) &   \textbf{95.21({$\uparrow$ 1.22})}    &  \textbf{77.61({$\uparrow$ 3.53})} \\
    \midrule
    ResNet362 &    93.45   & 70.41 \\
    ResNet362+SEM(ours) &   \textbf{95.45({$\uparrow$ 2.00})}    &  \textbf{77.72({$\uparrow$ 7.31})} \\
    \bottomrule
    \end{tabular}%
  \label{tab:depth}%
\end{table}%

\noindent\textbf{Depth of Backbone Models. }
Using ResNet47, ResNet164, ResNet272, and ResNet362 as backbone models, we compare our SEM with ResNet to dig out the effectiveness of the depth of backnone models. As shown in Table~\ref{tab:depth}, SEM can improve the performance of backbone models of each depth by 0.06\% to 7.31\%. Besides, we can learn from the experiment results that as the depth increasing, the performances of ResNet is not promoted consistently. This phenomenon may be related to the gradient dispersion of deep neural networks. However, the performances of SEM is consistent of the increasing of the depth, which means that our SEM can 
enhance the backbone models.

\section{Ablation Study}

\noindent\textbf{The Size of $\text{EO}$. }We discuss the size of $\text{EO}$ to examine whether it is necessary to integrate excitation operators. The experiment results is displayed in Table~\ref{tab:size}, showing that integrating excitation operators is effective. Although the performance of each operator is different, when the size of $\text{EO}$ increases, the performance of SEM consistently increases. This experiment shows that the size of $\text{EO}$ is positively related to the performance of SEM, and the switching strategy we proposed is stable and effective.
\begin{table}[htbp]
  \centering
  \caption{Comparison of different size of $\text{EO}$ on CIFAR10 and CIFAR100 in terms of Top-1 accuracy (in \%). }
    \begin{tabular}{cccc}
    \toprule
    The Size of $\text{EO}$ & Module & CIFAR-10  & CIFAR-100 \\
    \midrule
    \multicolumn{1}{c}{\multirow{3}[2]{*}{N=1}} & FC    & 94.32 & 75.28 \\
          & CNN   & 94.51 & 75.39 \\
          & IE    & 94.44 & 75.92 \\
    \midrule
    \multirow{3}[2]{*}{N=2} & FC, CNN &   94.92    & 76.37 \\
          & FC, IE &   94.80    & 76.52 \\
          & CNN, IE &   94.81    & 76.48 \\
    \midrule
    N=3   & FC, CNN, IE & 94.95 & 76.76 \\
    \bottomrule
    \end{tabular}%
  \label{tab:size}%
\end{table}%

\noindent\textbf{Removal of Decision Module. }We further verify the effectiveness of decision module. We experiment without decision weight which means that in Eq.(\ref{eq:switch}), $w_{\ast}$ is set as 1. The results is 94.72\% and 76.29\% on CIFAR10 and CIFAR100 respectively, which is 0.23\% and 0.47\% lower than SEM respectively in terms of Top-1 accuracy. These results verify that $w_{\ast}$ is effective and our decision module has good recognition ability for various excitation operators.

\noindent\textbf{Activation Funtion. }We explore the effects of different activation functions on Eq.(\ref{eq:switch}). According to the experiment results in Table~\ref{tab:activation}, Eq.(\ref{eq:switch}) is sensitive on the choice of activation functions, and Sigmoid is the optimal activation function. In particular, linear activation functions Relu and LeakyRelu are not suitable for Eq.(\ref{eq:switch}), while the nonlinear activation functions Tanh and Sigmoid perform relatively well, indicating that the symmetric activation function is more suitable for SEM.

\begin{table}[htbp]
  \centering
  \caption{Comparison of different activation functions on CIFAR100 in terms of Top-1 accuracy (in \%).}
    \begin{tabular}{lc}
    \toprule
    Activation Function & CIFAR-100 \\
    \midrule
    Tanh  & 70.11 \\
    ReLU  & 38.98 \\
    LeakyReLU & 39.32 \\
    \textbf{Sigmoid} & \textbf{76.76} \\
    \bottomrule
    \end{tabular}%
  \label{tab:activation}%
\end{table}%

\noindent\textbf{Without Data Augment. }To verify the ability of SEM to reduce overfitting, We train the models without data augment to reduce the influence of regularization from data augment. As shown in Table~\ref{tab:data-aug}, SEM achieves lower testing error than ResNet164 and SENet. To some extent, the switchable structure of SEM may have regularization effect.

\begin{table}[htbp]
  \centering
  \caption{Comparison of the impact of data augment on different attention modules on CIFAR10 and CIFAR100 in terms of Top-1 accuracy (in \%).}
    \begin{tabular}{l|cc}
    \toprule
    Model & CIFAR-10  & CIFAR-100 \\
    \midrule
    \multicolumn{1}{p{10em}|}{ResNet164 } & 87.18 & 60.82 \\
    SENet  & 88.24 & 62.87 \\
    SEM   & \textbf{89.58} & \textbf{67.27} \\
    \bottomrule
    \end{tabular}%
  \label{tab:data-aug}%
\end{table}%

\section{Conclusion}
In this paper, we focus on learning effective channel attention for deep CNNs with switchable mechanism. To this end, we propose an switchable excitation module (SEM), which can automatically decide to select and integrate attention operators to compute attention maps. Experimental results demonstrate our SEM is an plug-and-play block to improve the performance of deep CNN architectures and may have regularization effect. Moreover, our SEM have stable performance on backbone networks of various depths. In future, we will apply our SEM to more CNN architectures and tasks to further investigate incorporation of SEM with self-attention modules.

%
%
%
\bibliographystyle{splncs04}
\bibliography{ref}

\end{document}